\documentclass[sigconf]{acmart}
\usepackage{algorithm}
\usepackage{algpseudocode}
\usepackage{adjustbox}
\usepackage{placeins}
\usepackage{multirow}
\usepackage{url}

\setcopyright{none}
\renewcommand\footnotetextcopyrightpermission[1]{}
\AtBeginDocument{%
  }

\begin{document}

\title{PMG: Progressive Motion Generation via Sparse Anchor Postures Curriculum Learning}


\author{Yingjie Xi}
\email{yxi@bournemouth.ac.uk}
\affiliation{%
  \institution{Bournemouth University}
  \city{Bournemouth}
  \state{Dorset}
  \country{UK}
}

\author{Jian Jun Zhang}
\email{jzhang@bournemouth.ac.uk}
\affiliation{%
  \institution{Bournemouth University}
  \city{Bournemouth}
  \state{Dorset}
  \country{UK}
}

\author{Xiaosong Yang}
\email{xyang@bournemouth.ac.uk}
\affiliation{%
  \institution{Bournemouth University}
  \city{Bournemouth}
  \state{Dorset}
  \country{UK}
}

% \affiliation{%

% \affiliation{%

% \affiliation{%

% \affiliation{%

% \affiliation{%


\begin{abstract}
In computer animation, game design, and human-computer interaction, synthesizing human motion that aligns with user intent remains a significant challenge. Existing methods have notable limitations: textual approaches offer high-level semantic guidance but struggle to describe complex actions accurately; trajectory-based techniques provide intuitive global motion direction yet often fall short in generating precise or customized character movements; and anchor poses-guided methods are typically confined to synthesize only simple motion patterns. To generate more controllable and precise human motions, we propose \textbf{ProMoGen (Progressive Motion Generation)}, a novel framework that integrates trajectory guidance with sparse anchor motion control. Global trajectories ensure consistency in spatial direction and displacement, while sparse anchor motions only deliver precise action guidance without displacement. This decoupling enables independent refinement of both aspects, resulting in a more controllable, high-fidelity, and sophisticated motion synthesis. ProMoGen supports both dual and single control paradigms within a unified training process. Moreover, we recognize that direct learning from sparse motions is inherently unstable, we introduce \textbf{SAP-CL (Sparse Anchor Posture Curriculum Learning)}, a curriculum learning strategy that progressively adjusts the number of anchors used for guidance, thereby enabling more precise and stable convergence. Extensive experiments demonstrate that ProMoGen excels in synthesizing vivid and diverse motions guided by predefined trajectory and arbitrary anchor frames. Our approach seamlessly integrates personalized motion with structured guidance, significantly outperforming state-of-the-art methods across multiple control scenarios. \color{blue}{\url{https://github.com/2022yingjie/ProMoGen.git}}
\end{abstract}



\begin{teaserfigure}
  \includegraphics[width=\textwidth]{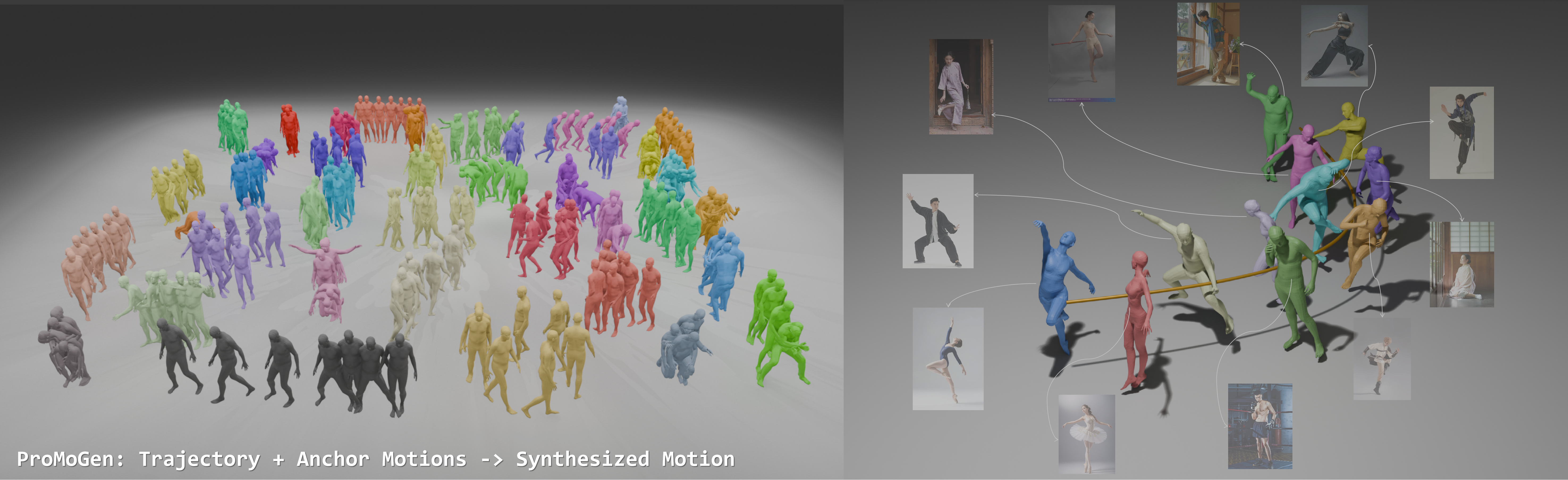}
  \caption{The left panel illustrates the synthesized results of the ProMoGen framework, under the control of both predefined trajectory and arbitrary anchor motions. The right panel delineates the inference process of our framework, wherein specific poses extracted from various images, as anchor motions, are combined with a predefined trajectory to synthesize realistic and customized human motion.}
  % \Description{Enjoying the baseball game from the third-base
  % seats. Ichiro Suzuki preparing to bat.}
  \label{fig:teaser}
\end{teaserfigure}


\maketitle

%%%===================================================================%%
\begin{figure*}[htbp]
  \centering
  \includegraphics[width=0.85\linewidth]{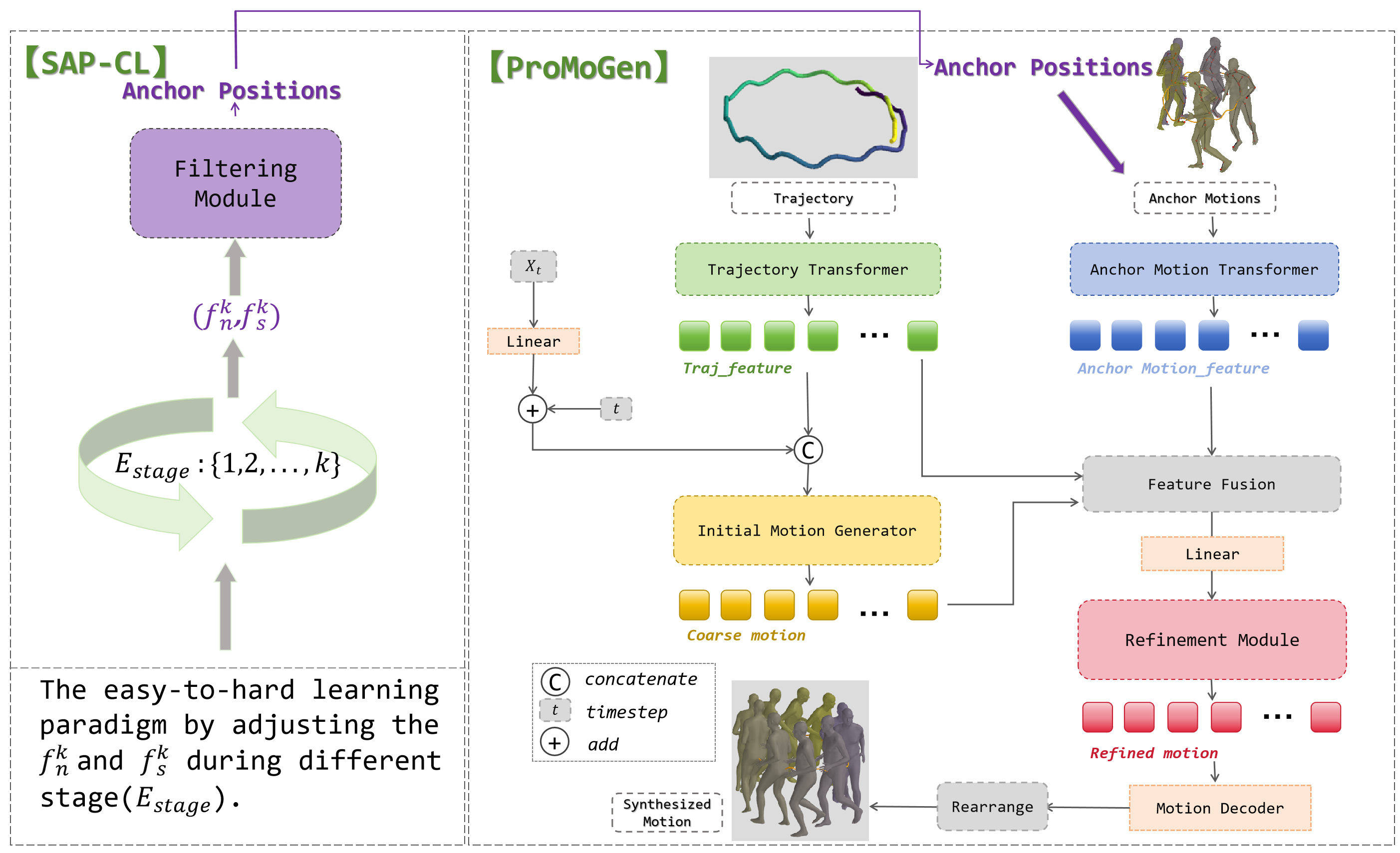}
  \caption{The SAP-CL training strategy(left panel) and the structure of ProMoGen method(right panel).}
  % \Description{A woman and a girl in white dresses sit in an open car.}
  \label{framework}
\end{figure*}
%%%===================================================================%%
\section{Introduction}
Human motion synthesis plays a crucial role in the field of cinematic character animation, game design, human-robot interaction, and virtual avatars. In these domains, there is a growing need for methods that can produce high-fidelity, expressive motion that is both semantically and environmentally aligned with specific thematic styles, while maintaining flexibility to accommodate diverse user requirements and interaction scenarios. Therefore, achieving precise control over motion trajectories and character postures is necessary to keep the generated motion realistic and desired. 

Traditional human motion synthesis methods primarily include text-to-motion~\cite{petrovich2022temos,wang2023t2m,zhang2023generating,lu2023humantomato,ma2024contact,dabral2023mofusion,xie2024omnicontrolcontroljointtime,tevet2022human,chen2024taming,huang2024stablemofusion,chen2024pay} synthesis, trajectory-to-motion synthesis~\cite{kania2021trajevae,wan2024tlcontrol,guo2025motionlab}, and audio-based synthesis~\cite{zhou2025exges,zhang2025motion,chhatre2024emotional,wan2024tlcontrol}, which have enabled coarse alignment between these conditions inputs and the generated motions. While these approaches are capable of generating smooth and simple human motion sequences, they still face significant challenges in producing highly customized or personalized complex motions that meet user-specific requirements~\cite{gang2025human}. For example, the specific posture at the selected position on the trajectory is the key to personalized motion expression, and this posture is often difficult to accurately describe using conventional text and other conditions. This limitation arises not only from the distribution of existing datasets, which tend to focus on simple motion patterns, but also from more fundamental issues, that the semantic ambiguity inherent in text descriptions and the geometric sparsity of trajectory signals, therefore further loss the interaction ability with the practical environment~\cite{cong2024laserhuman,ma2024contact,qu2024gpt}. These factors often fail to provide sufficient guidance for specifying fine-grained motion details, hindering the generation of more complex and personalized motion sequences. Other works~\cite{kim2022conditional,IEEE} adopts transformer-based end-to-end network to make an interpolation task for motion in-between,but directly end-to-end learning paradigm cannot learn motion distribution well, resulting low-fidelity and single synthesis. The~\cite{karunratanakul2023guided} explores the interplay between text and trajectories, as well as text and anchor posetures positions, but it did not effectively decouple these features nor achieve refined control over human motion generation.  

To address these limitations, we propose ProMoGen(Progressive Motion Generation), a diffusion-based framework~\cite{ho2020denoising, song2020denoising,ho2022video,yang2023diffusion}. By synergizing trajectory control with sparse anchor motions guidance, our method facilitates hierarchically controllable, customized, and precise motion synthesis. The key insight driving our approach is the decomposition of motion generation into two complementary subspaces: a global trajectory that governs macroscopic movement patterns, and local poses that represent stylistic content, towards to more diverse and free motion combination with control of arbitrary trajectory or motions. Specifically, to ensure robust understanding of global trajectories, we design the global Trajectory Feature Encoder (TFE), a transformer-based~\cite{peebles2023scalable,vaswani2017attention} architecture that projects user-defined trajectories into a latent space, capturing both temporal displacement and directional coherence. In parallel, to dynamically guide motion synthesis under varying key postures, we develop the anchor Filtering Module, which samples sparse anchor poses based on predefined temporal density and interval elasticity to get specific anchor pose controls. Then an Anchor Motion Encoder (AME) is introduced to learn features from the sparse pose conditions. To effectively integrate the outputs of these components, we design a Initial Motion Generator and Refinement Module, the former is responsible to build coarse motion under the control of trajectory only, to roughly align the latent representations of postures and trajectory, and the latter is to ensure smooth and high-fidelity interpolation while maintaining input pose original details. Besides, the diffusion process significantly enhances the synthesis capabilities of ProMoGen, enabling the generation of realistic and precisely controlled motions.

In addition, traditional training directly using sparse signals(e.g., less than 5\% frame density) often results in unstable gradient propagation due to insufficient supervisory information, hindering effective learning. Moreover, models trained under the specific conditions may not generalize well to varying inference scenarios. To alleviate this challenge, we introduce a novel training strategy, termed Sparse Anchor Posture Curriculum Learning (SAP-CL)~\cite{bengio2009curriculum,soviany2021curriculum,zhang2021flexmatch,zhang2019curriculum,zhao2021automatic,kim2024strategic,muttaqien2024mobile,manela2022curriculum,liu2022competence,wang2024grounded}. Intuitively, in this work, the tasks with denser anchor postures are simpler, enabling model easily handle the task by providing more informative supervisory signals, then gradually increase the difficulty level to get a better convergence. Following this easy-to-hard paradigm, our approach begins by utilizing denser anchor posture guidance. This allows the model to learn fundamental motion features with more supervision. Subsequently, we progressively reduce the density of anchor poses, enabling the model to adapt to more challenging, sparse conditions while maintaining generalization. 
% The experiments demonstrates that our SAP-CL fosters efficient knowledge transfer across varying keyframe postures complexities by ensuring stable convergence in every stages and enhancing the model's ability to synthesize high-fidelity motions under sparse conditions, ultimately leading to more precise and adaptable motion synthesis.

We conduct extensive experiments to evaluate the performance of ProMoGen across various metrics. The results demonstrate that ProMoGen produces smooth and visually appealing results, outperforming existing methods.

The main contributions of this work are as follows:
\begin{itemize}
    \item We propose a multi-guidance task that is used to synthesize human motion under the control of global trajectory and local postures, exploring the interactions between these two signals.
    \item We introduce ProMoGen, a well-designed network that integrates trajectory guidance with sparse anchor postures, enabling the generation of more controllable, high-fidelity.
    \item We introduce SAP-CL(Sparse Anchor Posture Curriculum Learning), this approach implements an easy-to-hard paradigm, ensuring stable convergence and facilitating motion synthesis in various scenarios by progressively increasing task complexity.
    \item We evaluate ProMoGen on multiple datasets and different scenarios, showcasing its superior performance compared to existing baselines. Ablation studies further validate the effectiveness of our proposed architecture and training strategy design.
\end{itemize}

%==================================Figure Sctter============================%
\begin{figure}
    \centering
    \includegraphics[width=0.8\linewidth]{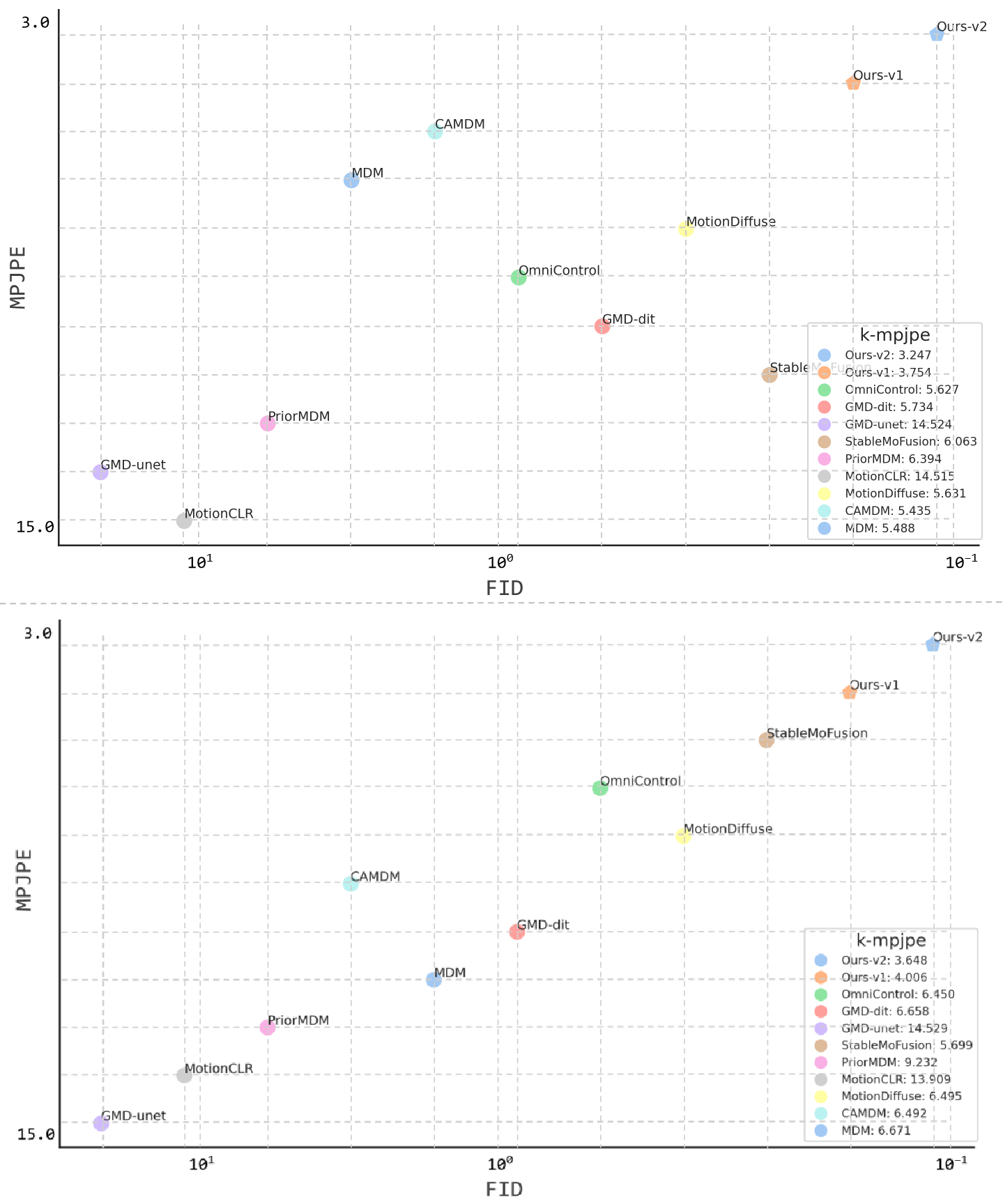}
    \caption{In this figure, the horizontal axis represents FID, while the vertical axis corresponds to MPJPE(To ensure comparability, the values of all models have been transformed to a uniform scale). Notably, data points that approach the upper right region along the diagonal indicate superior model performance. The v1 model achieves the most favorable results, thereby substantiating the robustness of our whole structure design. Additionally, the placement of v2 in the upper right corner further demonstrates the exceptional performance of our complete model with fine-designed components.}
    \label{fig:fid_scatter}
    \vspace{-15pt}
\end{figure}
%==========================================================================%

%==================================Relatd Work============================%
\section{Related Work}
\subsection{Motion Generation}
With Variational Autoencoders(VAEs)~\cite{guo2024momask,tevet2022motionclip,ma2024contact,lu2023humantomato,zhang2023generating,wang2023t2m,petrovich2022temos} and Diffusion models~\cite{chen2024pay,huang2024stablemofusion,chen2024taming,tevet2022human,xie2024omnicontrolcontroljointtime,dabral2023mofusion,chi2024m2d2m,athanasiou2024motionfix} emerging, human motion generation has evolved significantly. Among these advanced works, the text-to-motion task is one of the most compelling research direction~\cite{zhu2023human}, it leverages natural language descriptions as prompt input to synthesize realistic and semantically aligned human motions. Despite these improvements, text-based methods still face challenges~\cite{wang2024move,wang2025fg,shan2024towards} in precisely controlling motion trajectories and fine-grained kinematic details. On the on hand, text-based methods often struggle with ambiguous~\cite{lee2025multilingual,kim2024aligning,kamath2024scope,keluskar2024llms}, leading to misaligned or overly generic motions. On the other hand, it is difficult to describe complex and delicate human movements without other prior input, let alone achieving highly personalized or customized motion synthesis. Trajectory based~\cite{zhang2023finemogen,shafir2023human,athanasiou2023sinc,dai2024motionlcm,guo2025motionlab,wan2024tlcontrol,kania2021trajevae} methods are proposed to focus on generating motions conditioned on predefined spatial paths, offering more explicit control over the global movement of characters. But they primarily focus on coarse path alignment rather than fine-grained pose control. Additionally, motion in-between~\cite{chu2024real,cohan2024flexible,dai2025synthesizededitablemotioninbetweening,jiang2025packdit} focuses on generating complete motion sequences given key-frame poses. 

We introduce ProMoGen, a novel framework for motion synthesis that synergistically exploits both trajectory information and anchor postures. By integrating these elements, ProMoGen offers explicit, robust global guidance alongside fine-grained local constraints, resulting in enhanced performance on complex, customized tasks. In contrast, prior approaches~\cite{kim2022conditional,karunratanakul2023guided} employ anchor conditioning through simple input masking or the provision of global positions. Such methods encourage models to learn merely an interpolation space rather than a deep understanding of the motion distribution and the intricate interplay between trajectories and postures. Our ProMoGen is diffusion-based, which could further dig the intricate connection between two conditions and have a better generation.

\subsection{Curriculum learning}
Directly learning from anchor postures of random arrangement is challenging due to the sparsity. A more effective strategy is to build a progressive paradigm, from easy to difficult. Curriculum Learning (CL)~\cite{bengio2009curriculum} has been proposed to implement this strategy in model training, it trains models by gradually introducing examples of increasing difficulty. CL has demonstrated its power in improving the generalization capacity and convergence rate of various models in a wide range of scenarios such as computer vision~\cite{soviany2021curriculum,zhang2021flexmatch}, natural language processing~\cite{zhang2019curriculum,zhao2021automatic,kim2024strategic}, robotics and medical applications~\cite{muttaqien2024mobile,manela2022curriculum,liu2022competence,wang2024grounded}. We introduce its thoughts into our work and design SAP-CL, an easy-to-hard paradigm to improve model convergence and resulting in high-quality and high-fidelity motion.

%==================================Figure vis============================%
\begin{figure*}
    \centering
    \includegraphics[width=1.0\linewidth]{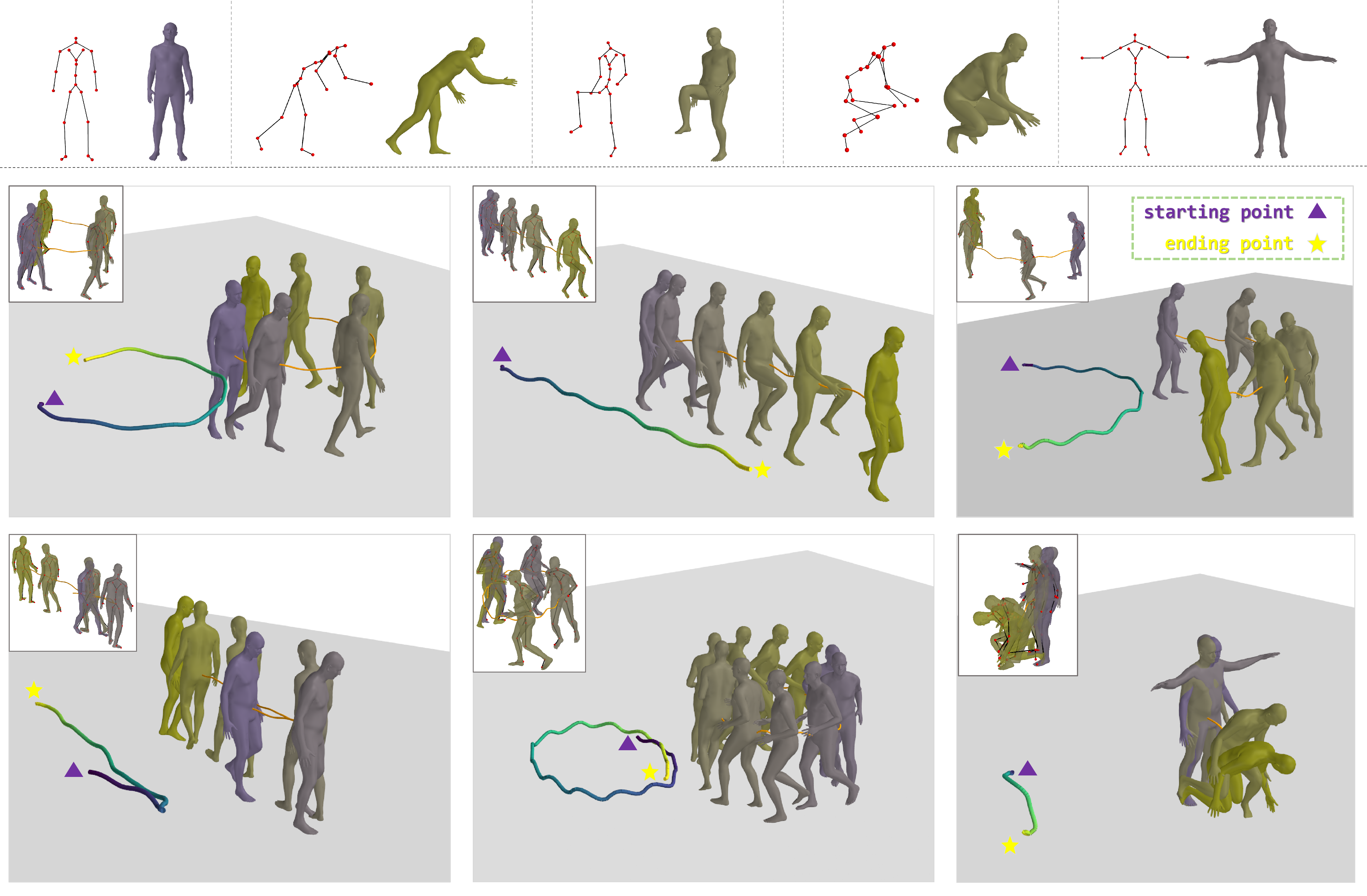}
    \caption{This figure illustrates the trajectory-based, sparse pose-guided motion generation process of our ProMoGen. In the top row, the reconstructed anchor motion (displayed as a mesh on the right) is compared with the control pose (depicted as a skeleton on the left), demonstrating a high degree of correspondence. In each of the subsequent sub-figures, the primary visualization represents the reconstructed motion, while the lines on the left indicate the trajectory. Color coding is applied such that both the character and the trajectory transition from purple (starting point) to yellow(ending point). The upper left corner of each sub-figure provides the guidance of sparse poses. All visualizations in this figure are derived from inferences made under the condition that the number of sparse anchor poses$f_n$ is fixed at five.}
    \label{fig:more_vis}
\end{figure*}
%==================================Figure===================================%

%========================================Method========================================%
\section{Method}
In this section, we first provide an overview of the diffusion approach that underpins our methodology. Next, we detail the design of our network architecture, finally followed by an explanation of the curriculum learning strategy employed in our framework. 

Let $X_0$ represents the original motion, the $\tau \in \mathbb{R}^{N \times 3}$ denote the predefined trajectory, where $N$ represents the length of the motion sequence and the value 3 corresponds to the global coordinates of the human pelvis at each frame. In addition, let $X_{s} \in \mathbb{R}^{M \times 6}$ denote the sparse anchor postures, where $M \in (0, N)$ represents the number of selected anchor postures, and the value 6 represents the motion feature. Our objective is to generate a plausible contiguous human motion sequence-$\hat{X}$-that exhibits coherent kinematic structures and smooth transitions.

\subsection{Preliminaries of Diffusion}
Diffusion models are generative frameworks that capture data distributions by progressively injecting noise into the data (the forward process) and subsequently learning to reverse this process via denoising (the reverse process). In our work, conditions provided include the trajectory \(\tau\), the sparse anchor postures \(X_s\), and the timestep \(t\), and our objective is to generate a complete motion sequence \(\hat{X}\).

Starting with an initial sample \(X_0\) drawn from the motion data distribution, the forward process iteratively adds Gaussian noise along a fixed Markov chain of \(T\) steps until the sample is transformed into pure Gaussian noise, i.e., \(X_T \sim \mathcal{N}(0, I)\). This process is formally defined as follows:
\[
q(X_t \mid X_{t-1}) = \mathcal{N}\left(X_t; \sqrt{\alpha_t}\,X_{t-1},\, (1 - \alpha_t)I\right),
\]
where \(\alpha_t \in (0, 1)\) denotes the predefined noise schedule, \(\mathcal{N}(\cdot)\) represents a Gaussian distribution, and \(I\) is the identity matrix.

The reverse process approximates the true posterior \(q(X_{t-1}\mid X_t)\) by learning to denoise the data using a neural network \(\epsilon_\theta\). Notably, inspired by prior work~\cite{tevet2022human}, our model is designed to directly predict the motion sequence rather than the noise. Moreover, while the original DDPM sampling process requires a large number of timesteps (e.g., \(T\approx1000\)) to produce high-quality samples, we accelerate sampling by employing DPM-Solver++, a high-order ODE solver tailored for diffusion models.

The reverse process is formulated as the following differential equation:
\[
\frac{dX_t}{dt} = f(t)X_t + g(t)\epsilon_\theta(X_t, t, \tau, X_s),
\]
where \(f(t)\) and \(g(t)\) are coefficients derived from the noise schedule, and the neural network \(\epsilon_\theta\) directly predicts the generated motion $\hat X$, conditioned on the trajectory \(\tau\), the sparse anchor poses \(X_s\), and the timestep \(t\).

DPM-Solver++ resolves this equation using a semi-linear multi-step method, thereby significantly reducing the number of sampling steps. The update rule is given by:
\[
X_{t_{n+1}} = X_{t_n} + \Delta t \sum_{k=0}^{K} \gamma_k \epsilon_\theta^{(k)}(X_{t_{n-k}}, t_{n-k}, \tau, X_s),
\]
where \(\Delta t\) is the adaptive step size, \(\gamma_k\) are the coefficients for the \(k\)-th order approximation, and \(\epsilon_\theta^{(k)}\) represents the \(k\)-th order derivative of the network prediction. This method enables high-quality sampling in as few as 20–30 steps, offering a substantial efficiency improvement over the original DDPM process.

\subsection{The structure of ProMoGen}
Before formal training, the process begins by setting a specific temporal density $f_n$ and interval elasticity $f_s$, then these two values are fed into the Filtering Module we designed to build the sparse anchor postures \(X_s\). Next, the trajectory \(\tau\) and the sparse postures \(X_s\) are fed into the ProMoGen to predict the synthesized motion \(\hat{X}\). We adopt Dit framework~\cite{ma2024latte,yang2024cogvideox}, to develop fine-designed encoders and a refinement module to enhance motion synthesis. An overview of all components is presented below, also shown in the right panel of the Figure \ref{framework}.

\subsubsection{\textbf{Filtering Module}}
In order to enable users to arbitrarily modify the insertion positions of postures, we introduce a Filtering Module (FM) that leverages a probabilistic algorithm during the training stage to imitate users' operations. This algorithm selects a set of anchor postures from a sequence of length \(N\) while satisfying two key constraints: (1) a fixed number of anchor poses, defined as the temporal density \(f_n\), and (2) a minimum interval between any two selected anchor frames, termed as interval elasticity \(f_s\). By uniformly sampling all potential valid combinations of anchor postures under these user-specified constraints, our design facilitates a flexible combination of conditions and enhances the system's adaptability, extremely extend the practical scale of training data.

The process is executed in three distinct steps: 
1. Virtual sampling point selection, 
2. Interval increment calculation, and 
3. Anchor position mapping, which collectively use the parameters \(f_n\) and \(f_s\).

Initially, \(f_n\) virtual positions are randomly selected from the set:
\[
\mathcal{P} = \{ p_1, p_2, \dots, p_{f_n} \} \subset \{ 1, 2, \dots, T_{\text{total}} \}, \quad |\mathcal{P}| = f_n,
\]
where the total number of virtual sampling points is given by \(T_{\text{total}} = R + f_n\). Here, 
\[
R = N - \left[ f_n + (f_n - 1)f_s \right]
\]
represents the difference between the total sequence length \(N\) and the minimal length required by the rigid constraint. The value \(R\) quantifies the available extra intervals, which are distributed as the elastic interval increments \(\{\delta_i\}_{i=0}^{f_n}\) under the constraint: $\sum_{i=0}^{f_n} \delta_i = R.$
These elastic interval increments correspond to the additional spacing beyond the predefined minimum \(f_s\) between consecutive anchors, and are defined as follows:
$$
\vspace{-5pt}
\begin{cases}
\delta_0 = p_1 - 1, \\
\delta_i = p_{i+1} - p_i - 1, & \text{for } 1 \leq i < f_n, \\
\delta_{f_n-1} = T_{\text{total}} - p_{f_n}.
\end{cases}
\vspace{-5pt}
$$
Finally, the anchor positions \(X_s = \{ x^1_s, x^2_s, \dots, x^{f_n}_s \}\) are determined by the mapping:
\[
\begin{cases}
x_1 = \delta_0 + 1, \\
x_j = x_{j-1} + f_s + \delta_{j-1} + 1, & \forall j > 1.
\end{cases}
\]
This recurrence can be expressed in closed form as:
$$
\vspace{-3pt}
x_j = j + (j-1)f_s + \sum_{i=0}^{j-1} \delta_i.
\vspace{-3pt}
$$
Regarding the properties of the algorithm, the interval between any two adjacent anchors is given by:
\[x_{k} - x_{k-1} = f_s + \delta_{k-1} + 1,\]
which guarantees that:
\[x_{k} - x_{k-1} \geq f_s + 1 > f_s.\]
Thus, the design robustly ensures that the minimum spacing condition is strictly maintained. During the training stage, the FM is both computationally efficient and mathematically rigorous to imitate the user-specific sparse anchor postures constraints.

%%=================================Figure comprison===================================%%
\begin{figure*}
    \centering
    \includegraphics[width=1.0\linewidth]{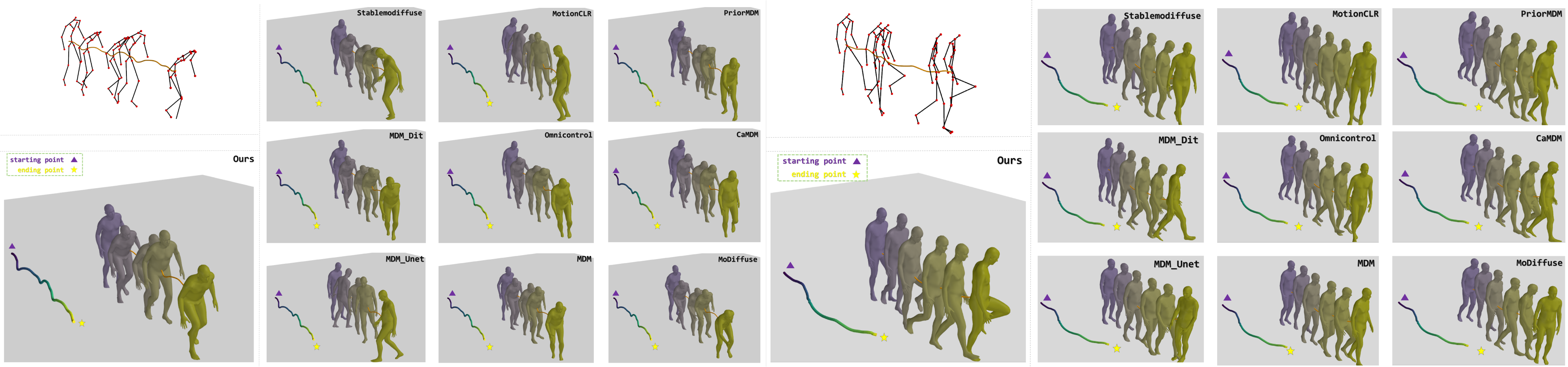}
    \caption{The figure illustrates the comparative generation effects of various modules. Under identical trajectory and anchor postures, the motions synthesized by our model are notably smoother and exhibit richer dynamic variations. Furthermore, the generated motions of ProMoGen most accurately adhere to the sparse poses guidance, highlighting the superior performance of our approach.}
    \label{fig:compare}
\end{figure*}

\begin{figure}
    \centering
    \includegraphics[width=1.0\linewidth]{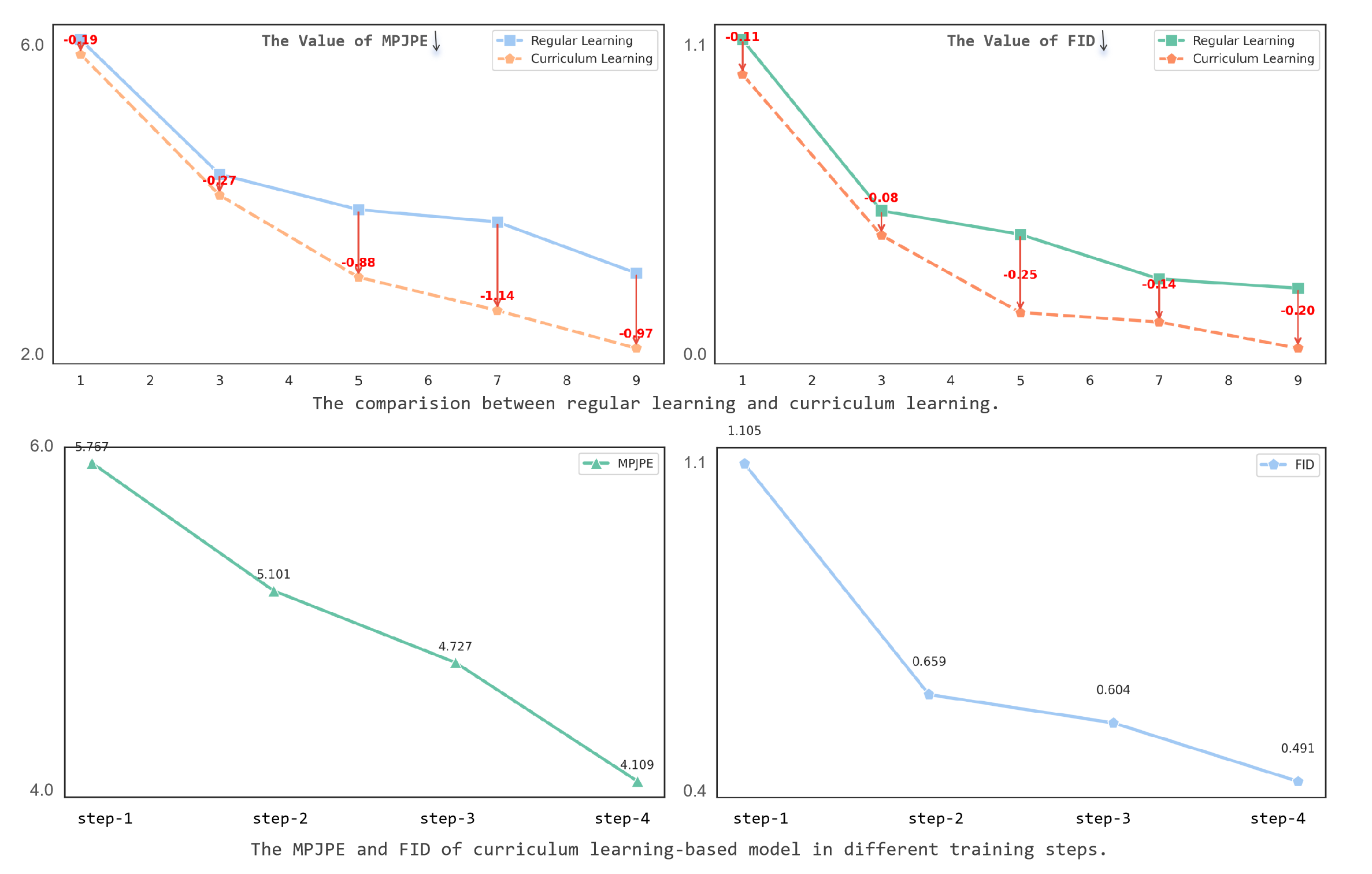}
    \caption{The two figures above display the changing of FID and MPJPE metrics, comparing regular training with curriculum learning. The curriculum learning yields significant improvements in model performance across varying numbers of anchor poses. The subfigures below further demonstrate that as the curriculum learning stages progress, model performance progressively enhances.}
    \label{fig:cur_result}
    \vspace{-3pt}
\end{figure}

\begin{figure}
    \centering
    \includegraphics[width=1.0\linewidth]{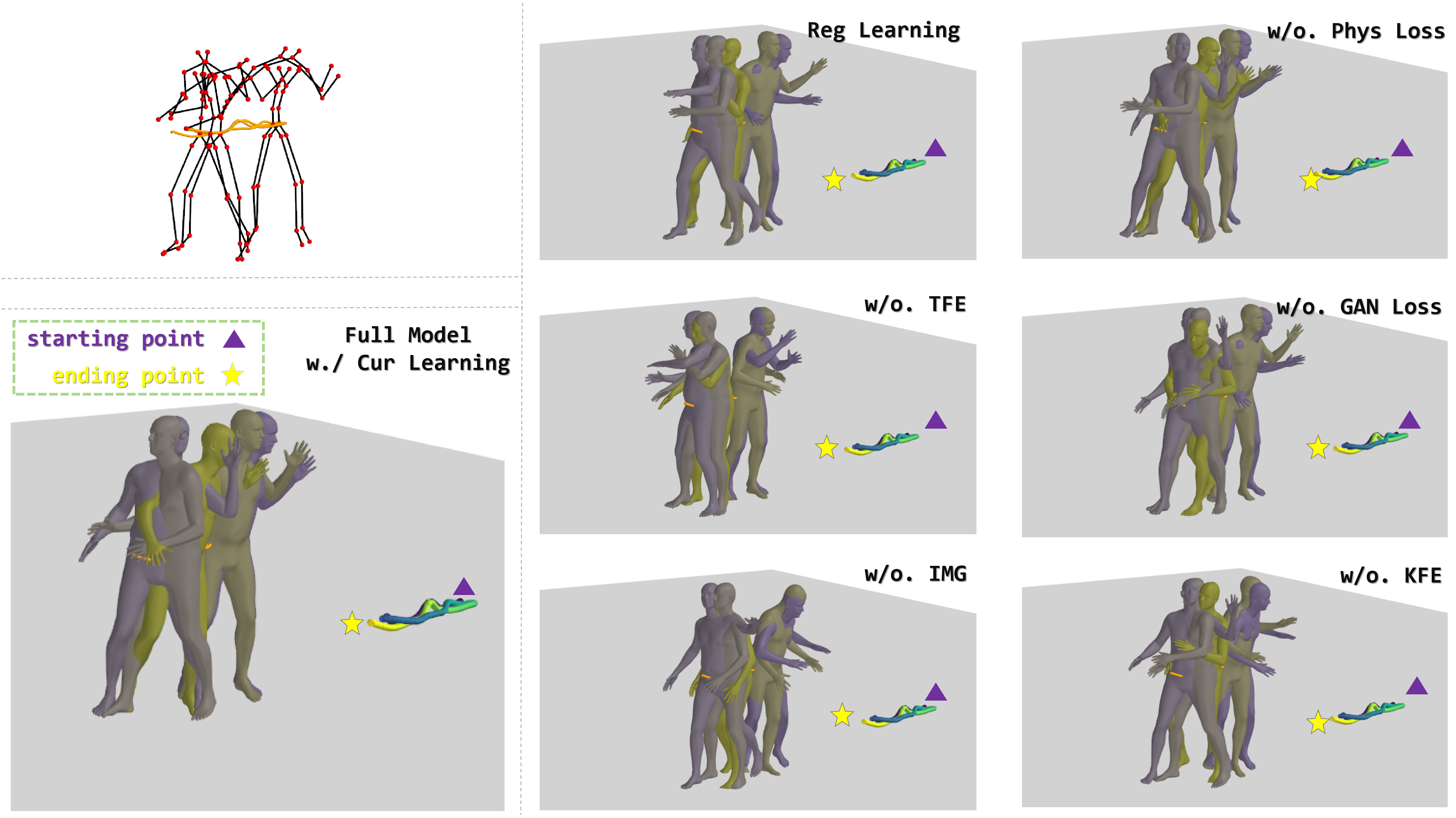}
    \caption{The figure presents a series of ablation experiments that examine the effects of various modules and loss functions on the overall performance of the model.}
    \label{fig:ablation}
    \vspace{-13pt}
\end{figure}
%%===================================================================================%%

%%=================================Network Design===================================%%
\subsubsection{\textbf{Network Design}}
The proposed network is a hierarchical transformer-based architecture designed conditioned human motion synthesis with the control of trajectory and sparse anchor motions guidance. It comprises five core modules:

1). Trajectory Encoder \(\mathcal{E}_{\tau}\): Maps the raw trajectory \(\tau\) into a latent space.
2). Anchor Motion Encoder \(\mathcal{E}_{k}\): Encodes the sparse postures constraints into contextual features.
3). Initial Motion Generator \(\mathcal{G}\): Synthesizes coarse motion priors conditioned solely on trajectory features, as defined by:
   \[
   M_{init} = \mathcal{G}(x_t, \mathcal{E}_{k}(X_s)) \in \mathbb{R}^{N \times 512}.
   \]
4). Refinement Module \(\mathcal{R}\): Jointly refines the motion priors by integrating both trajectory and anchor pose features:
   \[
   M_{refined} = \mathcal{R}(M_{init} \oplus \mathcal{E}_{k}(X_s) \oplus \mathcal{E}_{\tau}(\tau)),
   \]
   where \(\oplus\) denotes feature concatenation.
5). Decoder \(\mathcal{D}\): Projects the refined features into full kinematic motion sequences.

All modules adopt Dit(Diffusion Transformer) architecture for effective spatial-temporal modeling. This architecture facilitates motion distribution learning of trajectory-aligned motions while ensuring the anchor poses constraints, and have a good Spatial-Temporal learning and prediction ability.

\subsubsection{\textbf{The optimization target}}
The overall loss function of our framework is composed of several components designed to enforce both global consistency and local accuracy in the synthesized motion. Specifically, the loss terms include: 1)\(L_2\) Loss: Directly computes the Euclidean difference between the ground truth motion \(X\) and the synthesized motion \(\hat{X}\); 2)Anchor Motion Loss \(\mathcal{L}_{anchor}\): Specifically penalizes discrepancies in the postures at selected anchor positions; 3)GAN Loss \(\mathcal{L}_G\): Enhances the naturalness and global coherence of the motion by leveraging adversarial training; 4)Joint Loss \(\mathcal{L}_J\): Ensures local accuracy by enforcing precise predictions for individual joints; 5)Physical Constraints \(\mathcal{L}_P\): Imposes realistic motion through constraints such as foot slip reduction, body float minimization, and ground penetration prevention.

The total loss is thus formulated as:
\[
\mathcal{L}_{total} = \lambda_1\mathcal{L}_2 + \lambda_2\mathcal{L}_{anchor} + \lambda_3\mathcal{L}_J + \lambda_4\mathcal{L}_G + \lambda_5\mathcal{L}_P.
\]
During training, the weighting factors are set as \(\lambda_1 = \lambda_2 = \lambda_3 = 1.0\) and \(\lambda_4 = \lambda_5 = 0.1\).

\begin{table*}[htbp]
  \centering
  \begin{adjustbox}{width=\textwidth}
  \begin{tabular}{c|cc|cc|cc|cc|cc|cc}
    \hline
    \multirow{2}{*}{Method} & \multicolumn{2}{c|}{MPJPE $\downarrow$} & \multicolumn{2}{c|}{JS $\downarrow$} & \multicolumn{2}{c|}{Diversity $\rightarrow$} & \multicolumn{2}{c|}{DM $\uparrow$} & \multicolumn{2}{c|}{K-MPJPE $\downarrow$} & \multicolumn{2}{c}{FID $\downarrow$} \\
    \cline{2-13}& DS-1 & DS-2 & DS-1 & DS-2 & DS-1 & DS-2 & DS-1 & DS-2 & DS-1 & DS-2 & DS-1 & DS-2\\
    \hline
    MDM & {5.510} {\footnotesize ±0.103} & {6.685} {\footnotesize ±0.088} & \underline{0.020} {\footnotesize ±0.001} & {0.039} {\footnotesize ±0.001} & {4.455} {\footnotesize ±0.020} & {4.744} {\footnotesize ±0.015} & \underline{4.011} {\footnotesize ±0.029} & {1.845} {\footnotesize ±0.049} & {5.488} {\footnotesize ±0.102} & {6.671} {\footnotesize ±0.092} & {1.093} {\footnotesize ±0.030} & {1.119} {\footnotesize ±0.039}\\ %\hline
    CAMDM & {5.452} {\footnotesize ±0.068} & {6.607} {\footnotesize ±0.065} & \textbf{0.019} {\footnotesize ±0.001} & {0.035} {\footnotesize ±0.001} & {4.476} {\footnotesize ±0.011} & {4.691} {\footnotesize ±0.014} & {3.978} {\footnotesize ±0.031} & {1.981} {\footnotesize ±0.036} & {5.435} {\footnotesize ±0.069} & {6.492} {\footnotesize ±0.061} & {1.071} {\footnotesize ±0.027} & {1.122} {\footnotesize ±0.035}\\
    MotionDiffuse & {5.646} {\footnotesize ±0.092} & {6.531} {\footnotesize ±0.089} & {0.066} {\footnotesize ±0.001} & {0.038} {\footnotesize ±0.001} & {4.504} {\footnotesize ±0.014} & {4.735} {\footnotesize ±0.011} & {3.249} {\footnotesize ±0.037} & {2.004} {\footnotesize ±0.031} & {5.631} {\footnotesize ±0.090} & {6.495} {\footnotesize ±0.086} & {0.735} {\footnotesize ±0.022} & {0.804} {\footnotesize ±0.024}\\
    MotionCLR & {14.676} {\footnotesize ±0.164} & {14.187} {\footnotesize ±0.099} & {0.035} {\footnotesize ±0.001} & {0.033} {\footnotesize ±0.001} & {4.410} {\footnotesize ±0.011} & {4.646} {\footnotesize ±0.009} & {2.678} {\footnotesize ±0.051} & {1.471} {\footnotesize ±0.049} & {14.515} {\footnotesize ±0.156} & {13.909} {\footnotesize ±0.101} & {11.811} {\footnotesize ±0.076} & {8.703} {\footnotesize ±0.115}\\
    PriorMDM & {6.403} {\footnotesize ±0.101} & {9.225} {\footnotesize ±0.094} & {0.021} {\footnotesize ±0.001} & {0.041} {\footnotesize ±0.001} & {4.432} {\footnotesize ±0.016} & {4.725} {\footnotesize ±0.009} & {3.952} {\footnotesize ±0.031} & {1.680} {\footnotesize ±0.043} & {6.394} {\footnotesize ±0.101} & {9.232} {\footnotesize ±0.098} & {1.386} {\footnotesize ±0.045} & {1.729} {\footnotesize ±0.051}\\ 
    StableMoFusion & {6.112} {\footnotesize ±0.104} & {5.768} {\footnotesize ±0.070} & {0.033} {\footnotesize ±0.001} & \underline{0.032} {\footnotesize ±0.001} & {4.517} {\footnotesize ±0.013} & \underline{4.754} {\footnotesize ±0.014} & {3.813} {\footnotesize ±0.028} & \underline{2.158} {\footnotesize ±0.038} & {6.063} {\footnotesize ±0.102} & {5.699} {\footnotesize ±0.071} & {0.420} {\footnotesize ±0.013} & {0.512} {\footnotesize ±0.019}\\
    GMD-unet & {14.633} {\footnotesize ±0.131} & {14.535} {\footnotesize ±0.088} & {0.042} {\footnotesize ±0.001} & {0.037} {\footnotesize ±0.001} & {4.486} {\footnotesize ±0.012} & {4.640} {\footnotesize ±0.007} & {2.562} {\footnotesize ±0.050} & {1.413} {\footnotesize ±0.033} & {14.524} {\footnotesize ±0.133} & {14.529} {\footnotesize ±0.088} & {12.029} {\footnotesize ±0.047} & {8.896} {\footnotesize ±0.094}\\
    GMD-dit & {5.763} {\footnotesize ±0.076} & {6.652} {\footnotesize ±0.103} & {0.033} {\footnotesize ±0.001} & {0.036} {\footnotesize ±0.001} & \underline{4.521} {\footnotesize ±0.018} & {4.736} {\footnotesize ±0.009} & {3.777} {\footnotesize ±0.030} & {1.922} {\footnotesize ±0.048} & {5.734} {\footnotesize ±0.076} & {6.658} {\footnotesize ±0.102} & {0.887} {\footnotesize ±0.029} & {0.916} {\footnotesize ±0.034}\\
    OmniControl & {5.647} {\footnotesize ±0.058} & {6.464} {\footnotesize ±0.097} & {0.027} {\footnotesize ±0.001} & {0.035} {\footnotesize ±0.001} & {4.478} {\footnotesize ±0.015} & {4.743} {\footnotesize ±0.011} & {3.960} {\footnotesize ±0.034} & {1.941} {\footnotesize ±0.049} & {5.627} {\footnotesize ±0.060} & {6.450} {\footnotesize ±0.101} & {0.943} {\footnotesize ±0.018} & {0.885} {\footnotesize ±0.022}\\
    \hline
    Ours-v1 & \underline{4.189} {\footnotesize ±0.077} & \underline{4.289} {\footnotesize ±0.055} & {0.032} {\footnotesize ±0.002} & {0.073} {\footnotesize ±0.001} & {4.516} {\footnotesize ±0.015} & {4.732} {\footnotesize ±0.013} & {3.866} {\footnotesize ±0.055} & {1.806} {\footnotesize ±0.050} & \underline{3.754} {\footnotesize ±0.061} & \underline{4.006} {\footnotesize ±0.055} & \underline{0.418} {\footnotesize ±0.012} & \underline{0.438} {\footnotesize ±0.020}\\
    Ours-v2 & \textbf{3.257} {\footnotesize ±0.062} & \textbf{3.654} {\footnotesize ±0.049} & {0.030} {\footnotesize ±0.001} & \textbf{0.032} {\footnotesize ±0.001} & \textbf{4.531} {\footnotesize ±0.015} & \textbf{4.768} {\footnotesize ±0.011} & \textbf{4.278} {\footnotesize ±0.041} & \textbf{2.656} {\footnotesize ±0.064} & \textbf{3.247} {\footnotesize ±0.062} & \textbf{3.648} {\footnotesize ±0.046} & \textbf{0.279} {\footnotesize ±0.012} & \textbf{0.412} {\footnotesize ±0.017}\\
    \hline
    % \hline
    % MRAA & 1.69E-04 & 29.72 & 0.597 & 0.397 & 257.16 & 136.13 & 1770.91 
  \end{tabular}
  \end{adjustbox}
  \caption{Performance comparison of various methods across multiple metrics on HumanML3D~\cite{Guo_2022_CVPR}(DS-1) and CombatMotion~\cite{CombatMotion}(DS-2) respectively. The best results are in bold, and the second best results are underlined. $\downarrow$ means the lower is better while $\uparrow$ means the higher is better. $\rightarrow$ represents the closer to the value of Real is better.}
  \label{tab:1}
  \vspace{-15pt}
\end{table*}

\subsection{Sparse Anchor Posture Curriculum Learning}
Under a total of \(E_{total}\) training epochs, the process is segmented into \(E_{stage}\) stages, each corresponding to a anchor sampling interval \(\left[K_{\min}^{(s)}, K_{\max}\right]\). Here, \(s \in \{1,\dots,E_{stage}\}\) denotes the stage index, and \(K_{\max}\) is fixed at 30, as we posit that incorporating more than 30 anchor motions will introduce redundant, homogeneous information. The initial stage employs the configuration with \(K_{\min}^{(1)} = 20\), and subsequent stages progressively reduce the minimum number of anchor postures, culminating in the final stage where a single action (\(K_{\min}^{(s)} = 1\)) is permitted.

During each stage $k$, the temporal density $f^{k}_n$ is constructed by randomly sampling \(f^{k}_n \in \left[K_{\min}^{(s)}, K_{\max}\right]\), the interval elasticity $f^{k}_s \in \left[4, \lfloor{\frac{N}{f^{k}_n}}\rfloor\right]$, which is used to set the minimum interval between any two anchor motions, the $f^{k}_s$ will not exceed the maximum value $\lfloor{\frac{N}{f^{k}_n}}\rfloor$. By gradually transitioning from dense (\(K_{\min} = 20\)) to sparse (\(K_{\min} = 1\)) configurations, the model incrementally adapts to the anchor-missing generation task, thereby mitigating the risk of mode collapse that can arise from direct sparse training. The whole process is shown in Algorithm ~\ref{alg:curriculum} 

\vspace{-5pt}
%%==========================================Table 2===================================%%
\begin{table*}[htbp]
    \centering
    \begin{adjustbox}{width=0.8\textwidth}
    \begin{tabular}{c|c|c|c|c|c|c|c}
    \hline
    Method & A-Num & {MPJPE $\downarrow$} & {JS $\downarrow$} & {Diversity $\rightarrow$} & {DM $\uparrow$} & {K-MPJPE $\downarrow$} & {FID $\downarrow$} \\
    \hline
    \multirow{10}{2.5cm}{Ours(Regular) \\vs. Ours(Curriculum)\centering} & 9-Reg & {3.101} {\footnotesize ±0.046} & {0.032} {\footnotesize ±0.001} & {4.504} {\footnotesize ±0.016} & {4.310} {\footnotesize ±0.025} & {3.099} {\footnotesize ±0.047} & {0.319} {\footnotesize ±0.010} \\
    \multirow{10}{*}{} & 9-Cur & \textbf{2.130} {\footnotesize ±0.029} & \textbf{0.031} {\footnotesize ±0.001} & \textbf{4.515} {\footnotesize ±0.012} & \textbf{4.506} {\footnotesize ±0.036} & \textbf{2.127} {\footnotesize ±0.028} & \textbf{0.124} {\footnotesize ±0.005} \\
    \cline{2-8}
    
    \multirow{10}{*}{} & 7-Reg & {3.762} {\footnotesize ±0.037} & {0.031} {\footnotesize ±0.001} & {4.505} {\footnotesize ±0.011} & {4.258} {\footnotesize ±0.027} & {3.234} {\footnotesize ±0.036} & {0.349} {\footnotesize ±0.013} \\
    \multirow{10}{*}{} & 7-Cur & \textbf{2.618} {\footnotesize ±0.051} & \textbf{0.030} {\footnotesize ±0.001} & \textbf{4.511} {\footnotesize ±0.015} & \textbf{4.403} {\footnotesize ±0.029} & \textbf{2.624} {\footnotesize ±0.052} & \textbf{0.209} {\footnotesize ±0.012} \\
    \cline{2-8}
    
    \multirow{10}{*}{} & 5-Reg & {3.922} {\footnotesize ±0.061} & {0.031} {\footnotesize ±0.001} & {4.493} {\footnotesize ±0.017} & {4.228} {\footnotesize ±0.043} & {3.911} {\footnotesize ±0.061} & {0.494} {\footnotesize ±0.019} \\
    \multirow{10}{*}{} & 5-Cur & \textbf{3.047} {\footnotesize ±0.040} & \textbf{0.030} {\footnotesize ±0.001} & \textbf{4.510} {\footnotesize ±0.012} & \textbf{4.285} {\footnotesize ±0.032} & \textbf{3.003} {\footnotesize ±0.039} & \textbf{0.240} {\footnotesize ±0.010} \\
    \cline{2-8}
    
    \multirow{10}{*}{} & 3-Reg & {4.379} {\footnotesize ±0.056} & {0.030} {\footnotesize ±0.000} & {4.483} {\footnotesize ±0.016} & {4.166} {\footnotesize ±0.024} & {4.341} {\footnotesize ±0.057} & {0.571} {\footnotesize ±0.015} \\
    \multirow{10}{*}{} & 3-Cur & \textbf{4.109} {\footnotesize ±0.070} & \textbf{0.029} {\footnotesize ±0.001} & \textbf{4.519} {\footnotesize ±0.014} & \textbf{4.212} {\footnotesize ±0.025} & \textbf{4.107} {\footnotesize ±0.069} & \textbf{0.491} {\footnotesize ±0.017} \\
    \cline{2-8}
    
    \multirow{10}{*}{} & 1-Reg & {6.119} {\footnotesize ±0.095} & {0.029} {\footnotesize ±0.001} & {4.482} {\footnotesize ±0.019} & \textbf{4.073} {\footnotesize ±0.034} & {6.077} {\footnotesize ±0.092} & {1.127} {\footnotesize ±0.033} \\
    \multirow{10}{*}{} & 1-Cur & \textbf{5.929} {\footnotesize ±0.084} & \textbf{0.028} {\footnotesize ±0.002} & \textbf{4.514} {\footnotesize ±0.011} & {4.055} {\footnotesize ±0.034} & \textbf{5.915} {\footnotesize ±0.086} & \textbf{1.014} {\footnotesize ±0.023} \\
    \hline
    \end{tabular}
    \end{adjustbox}
    \caption{The performance comparison of ProMoGen under both regular and curriculum learning paradigms-using varying numbers of anchor guidance along the same trajectory on the HumanML3D~\cite{Guo_2022_CVPR} dataset-is presented, with the best results highlighted in bold.}
    \label{tab:2}
    \vspace{-15pt}
\end{table*}
%%==========================================Table 2===================================%%

\begin{algorithm}[h]
\caption{Sparse Anchor Progressive Curriculum Learning}
\label{alg:curriculum}
    \begin{algorithmic}[1]
        \Require Total epochs $E_{\text{total}}$, predefined curriculum learning stages $E_{\text{stage}}$, maximum anchor poses $K_{\max}=30$
        \Ensure Trained diffusion model
            \State Initialize model parameters $\theta$
            \For{$s = 1$ \textbf{to} $E_{\text{stage}}$}
                \State \textbf{Compute minimum anchor poses:} 
                \If{$s = 1$}
                    \State $K_{\min}^{(s)} \gets 20$ \hfill $\triangleright$ Initial dense phase
                \ElsIf{$s = E_{\text{stage}}$}
                    \State $K_{\min}^{(s)} \gets 1$ \hfill $\triangleright$ Final sparse phase
                \Else
                    \State $K_{\min}^{(s)} \gets \left\lfloor 20 - \frac{19(s-1)}{E_{\text{stage}}-1} \right\rfloor$ \hfill $\triangleright$ Linear interpolation
                \EndIf
                \State \textbf{Stage epochs:} $E_{\text{stage\_epochs}} \gets E_{\text{total}} / E_{\text{stage}}$
                \For{$e = 1$ \textbf{to} $E_{\text{stage\_epochs}}$}
                    \State \textbf{Set temporal density and interval elasticity:} 
                    $f^{k}_n \sim \mathcal{U}(K_{\min}^{(s)}, K_{\max})$, 
                    $f^{k}_s \sim \mathcal{U}(4, \lfloor{\frac{N}{f^{k}_n}}\rfloor)$
                    \State \textbf{Build Anchor conditions:} 
                    $X_s \gets FM(f^{k}_n, f^{k}_s) \gets f^{k}_n, f^{k}_s$
                    \State \textbf{Train model:} $\theta \gets \theta - \nabla_\theta \mathcal{L}_{total}(x_t, t, \tau, X_s)$ \hfill $\triangleright$ Update via $\mathcal{L}_{total}$
                \EndFor
            \EndFor
            \State \Return $\theta$
    \end{algorithmic}
    % \vspace{-10pt}
\end{algorithm}
%Moreover, the stochastic selection of \(K_s\) within each stage, as opposed to a fixed value, compels the model to handle situations of varying number and positions of keyframes, ultimately enhancing its generalization capabilities.
%

%%==========================================Table 3===================================%%
\begin{table*}[htbp]
    \centering
    \begin{adjustbox}{width=0.7\textwidth}
    \begin{tabular}{c|c|c|c|c|c|c|c}
    \hline
    Method & Components & {MPJPE $\downarrow$} & {JS $\downarrow$} & {Diversity $\rightarrow$} & {DM $\uparrow$} & {K-MPJPE $\downarrow$} & {FID $\downarrow$} \\
    \hline
    \multirow{7}{2.5cm}{Ours Full Model\centering} & w/o. IMG & {3.670} {\footnotesize ±0.079} & {0.029} {\footnotesize ±0.001} & {4.499} {\footnotesize ±0.019} & {4.243} {\footnotesize ±0.030} & {3.672} {\footnotesize ±0.082} & {0.419} {\footnotesize ±0.016} \\
    \multirow{7}{*}{} & w/o. TFE & {7.202} {\footnotesize ±0.102} & \textbf{0.025} {\footnotesize ±0.001} & {4.452} {\footnotesize ±0.011} & {4.009} {\footnotesize ±0.044} & {7.193} {\footnotesize ±0.103} & {1.596} {\footnotesize ±0.041} \\
    % \cline{2-8}
    \multirow{7}{*}{} & w/o. AME & {3.531} {\footnotesize 0.066} & {0.032} {\footnotesize ±0.001} & {4.499} {\footnotesize ±0.017} & {4.264} {\footnotesize ±0.031} & {3.529} {\footnotesize ±0.065} & {0.371} {\footnotesize ±0.013} \\
    \cline{2-8}
    \multirow{7}{*}{} & w/o. GAN loss & {3.326} {\footnotesize 0.054} & {0.031} {\footnotesize ±0.001} & {4.495} {\footnotesize ±0.016} & {4.276} {\footnotesize ±0.036} & {3.321} {\footnotesize ±0.055} & {0.330} {\footnotesize ±0.014} \\
    \multirow{7}{*}{} & w/o. Phys loss & {3.251} {\footnotesize 0.046} & {0.030} {\footnotesize ±0.001} & {4.510} {\footnotesize ±0.015} & {4.265} {\footnotesize ±0.034} & {3.221} {\footnotesize ±0.045} & {0.296} {\footnotesize ±0.015} \\
    \cline{2-8}
    \multirow{7}{*}{} & w/. Reg Learning & {3.257} {\footnotesize ±0.062} & {0.030} {\footnotesize ±0.001} & \textbf{4.531} {\footnotesize ±0.015} & {4.278} {\footnotesize ±0.041} & {3.247} {\footnotesize ±0.062} & {0.279} {\footnotesize ±0.012} \\
    \multirow{7}{*}{} & w/. Cur Learning & \textbf{3.047} {\footnotesize ±0.040} & {0.030} {\footnotesize ±0.001} & {4.510} {\footnotesize ±0.012} & \textbf{4.285} {\footnotesize ±0.032} & \textbf{3.003} {\footnotesize ±0.039} & \textbf{0.240} {\footnotesize ±0.010} \\
    
    \hline
    \end{tabular}
    \end{adjustbox}
    \caption{Ablation studies for our model on HumanML3D~\cite{Guo_2022_CVPR} dataset, the best results are in bold.}
    \label{tab:3}
    \vspace{-15pt}
\end{table*}
%%==========================================Table 3===================================%%

%%=====================================Experiments=================================%%
\section{Experiments}
\textbf{Datasets}: We evaluate our method on two datasets. For quantitative comparisons between our framework and other approaches, we employ the HumanML3D~\cite{Guo_2022_CVPR} dataset, which consists of 14,646 motion sequences, and the CombatMotion~\cite{CombatMotion} dataset, which includes 14,883 motion sequences. To assess the impact of the curriculum learning strategy, we perform comparisons using the HumanML3D dataset. Additionally, ablation studies investigating the effects of various components are conducted on the HumanML3D dataset.

\textbf{Evaluation Metrics} Our framework is assessed using several metrics:
  1. MPJPE (Mean Per Joint Position Error): Computes the Euclidean distance between corresponding joints of the generated and ground truth motions.
  2. JS (Joint Smoothness): Evaluates the smoothness of transitions between adjacent frames.
  3. Diversity: Measures the variance across generated motions to assess variety.
  4. Directional Consistency: Quantifies the similarity between the predicted motion direction and the actual motion trajectory.
  5. K-MPJPE: Assesses the Euclidean distance specifically at anchor positions.
  6. FID (Fréchet Inception Distance)~\cite{heusel2017gans}: Evaluates the distributional gap between the generated and real motion data.

\textbf{Implement Details}: Because this work is to synthesize human motion by simultaneously conditioning on trajectory data and sparse anchor postures. In contrast, prior methods predominantly rely on text-based input. For fair comparison and to demonstrate the efficacy of our framework, we reconstruct these baseline methods by removing their text encoders and substituting them with two linear layers that serve as trajectory and sparse posture encoders, respectively. In our evaluations, this same variant is denoted as Ours-v1, while the full model incorporating all proposed components is as Ours-v2.

In order to fairly compare our model with other models, motions from all datasets have been retargeted into one skeleton following HumanML3D format with 20 fps, where the number of joint $\mathcal{J}$ is 22. The learning rate is set to be $0.0001$. The $E_{total}$ is set to be 100. The $E_{stage}$ is set as $4$. The timesteps are set to $1000$ for training and 25 for inference. Our models are trained on one RTX 4070Ti with each batch of 32. 

\vspace{-5pt}
\subsection{Quantitative and Qualitative Results}
As shown in Figure~\ref{fig:more_vis}, our framework exhibits robust capabilities in generating motion that is both trajectory-aligned and anchor poses aligned, resulting in the stylistic and dynamic consistent motion synthesis, thereby demonstrating its versatility and performance. Table~\ref{tab:1} illustrates that our ProMoGen variant utilizing a simple linear encoder (Ours-v1) already outperforms competing methods on most evaluation metrics, while the full ProMoGen model (Ours-v2) further make progress on these metrics. All experiments reported in Table~\ref{tab:1} were conducted under standard training conditions, thereby underscoring the superiority of our structure. 
%As evidenced by the results in Table~\ref{tab:1}, our ProMoGen model consistently achieves superior metrics compared to competing methods. 
The reason is our ProMoGen decouples trajectory and motion by assigning distinct encoders to each, thereby enhancing the understanding and mapping of their specific conditional features. In addition, our initial motion generator leverages trajectory information to produce a coarse global motion, establishing a rough alignment between trajectory and motion features, where the coarse motion is then progressively refined and focus exclusively on local details by the refinement module. This approach still embodies an easy-to-hard learning paradigm,ultimately leading to improved performance.

Figure~\ref{fig:compare} presents the comparative results, while Figure~\ref{fig:fid_scatter} illustrates the FID scatter plot for all methods listed in Table~\ref{tab:1}. To ensure the visual effect, the values for all models have been normalized to a uniform scale. Notably, our ProMoGen model is positioned in the upper right quadrant of the scatter plot, demonstrating its superior performance on both the HumanML3D~\cite{Guo_2022_CVPR} and CombatMotion~\cite{CombatMotion} datasets.

\vspace{-5pt}
\subsection{Curriculum Learning Studies}
Table~\ref{tab:2} demonstrates the effectiveness of our curriculum training strategy. The "A-Num" denotes the number of anchor postures used as conditioning inputs. As observed in Table~\ref{tab:2}, increasing the number of anchor poses generally leads to improved metric values, which is an intuitive outcome. Moreover, curriculum training yields significant improvements across nearly all metrics for all configurations. The upper section of Figure~\ref{fig:cur_result} provides a detailed comparison between the two training paradigms under identical conditions, while the lower section illustrates the reduction in both MPJPE and FID when employing a progressively easy-to-hard training approach.
% Table~\ref{tab:2} illustrates the effectiveness of our curriculum training strategy. The Kf-Num represents how many kerframe postures are leveraged as conditions. From the table~\ref{tab:2} it is obvious that the more key frames bring higher metric values, which is intuitive. And after the curriculum training, all situations get significant improvements on almost all metric values. The Figure 4 upper part shows the detailed change between these two training on the same conditions, and the lower part shows how the MPJPE and FID decrease by adopting progressively easy-to-hard paradigm.

\vspace{-5pt}
\subsection{Ablation Studies}
In this section, we conduct a series of ablation experiments to rigorously evaluate the contributions of individual components within our framework, as oberseved in Table ~\ref{tab:3}. Specifically, we sequentially remove the Initial Motion Generator (IMG), the Trajectory Feature Encoder (TFE), the Anchor Motion Encoder (AME) and two losses in separate experimental settings to assess their impact on overall performance.

The results demonstrate that the TFE plays a pivotal role in enhancing the model's capability, followed by IMG and AME, moreover, GAN and Phys Loss can further improve the precision of the results. Figure~\ref{fig:ablation} provides visualizations under various ablation conditions, clearly illustrating that our complete model reconstructs postures with high fidelity. These findings substantiate the critical contributions of each module and validate the robustness of our design.
% In this section, we perform several ablation experiments on our framework to validate the designs. We remove the IMG(Initial Motion Generator),TFE(Trajectory Feature Encoder),KFE(Keyframe Feature Encoder) respectively in the first three line. Then we compare the situations withour GAN loss and physical loss. The results show the TFE significantly influence the model ability. In addtion, in Figure 5, we visualized the different situation motion to demonstrate that our model rebuild the most similar postures and fluent motions.

\section{Conclusion}
Building on our proposed Trajectory-Sparse Postures-Motion task, we have developed the ProMoGen framework to synthesize human motion with high fidelity and flexibility. To accelerate the inference process, we have incorporated DPM-Solver++, a high-order ODE solver tailored for diffusion models. During the training, we design the Filtering Module to sample sparse anchor motions in uniformly probabilistic sampling from the predefined $(f_n, f_s)$, to extend the practical training scale and imitate the user-specific choice. Then we introduce a novel curriculum learning strategy SAP-CL that follows an easy-to-hard paradigm, which substantially enhances the quality of the generated motion. The extensive comparison experiments and ablations demonstrate the ability of ProMoGen and the effectiveness of SAP-CL. More details and results are shown in the \textbf{Appendix}.

In all, Our framework is capable of seamlessly fusing arbitrary motion cues by conditioning on a specified trajectory, thereby achieving smooth and highly flexible motion generation. Overall, the ProMoGen framework demonstrates superior versatility, performance, and efficiency when compared to existing state-of-the-art methods, underscoring its potential for advancing the field of human motion synthesis. 

% \newpage
\bibliographystyle{ACM-Reference-Format}
\bibliography{sample-base}

\end{document}